\documentclass{article}

\PassOptionsToPackage{square,numbers}{natbib}
  \usepackage[preprint]{neurips_2020}

\usepackage[utf8]{inputenc} 
\usepackage[T1]{fontenc}    
\usepackage{hyperref}       
\usepackage{url}            
\usepackage{booktabs}       
\usepackage{amsfonts}       
\usepackage{nicefrac}       
\usepackage{microtype}      

\usepackage{amsmath}
\usepackage{amsthm}
\usepackage{amsfonts}
\newtheorem{theorem}{Theorem}

\newtheorem{proposition}{Proposition}

\usepackage{graphicx}
\usepackage[labelformat=simple]{subcaption}
\usepackage{algpseudocode}
\usepackage[ruled,linesnumbered,noline]{algorithm2e}

\usepackage{pdfpages}

\newlength\myindent
\title{State Action Separable Reinforcement Learning}

\author{%
  Ziyao~Zhang \\
  Imperial College London\\
  London, United Kingdom \\
  \texttt{ziyao.zhang15@imperial.ac.uk} \\
   \And
   Liang~Ma \\
   IBM T. J. Watson Research Center \\
   Yorktown Heights, NY, United States \\
   \texttt{lianglondon@gmail.com} \\
   \AND
   Kin K.~Leung \\
   Imperial College London \\
   London, United Kingdom \\
   \texttt{kin.leung@imperial.ac.uk} \\
   \And
   Konstantinos~Poularakis \\
   Yale University \\
   New Haven, CT, United States \\
   \texttt{konstantinos.poularakis@yale.edu} \\
   \And
   Mudhakar~Srivatsa \\
   IBM T. J. Watson Research Center\\
   Yorktown Heights, NY, United States\\
   \texttt{msrivats@us.ibm.com} \\
}

\begin{document}

\maketitle

\begin{abstract}
Reinforcement Learning (RL) based methods have seen their paramount successes in solving  serial decision-making and control problems in recent years. For conventional RL formulations, Markov Decision Process (MDP) and state-action-value function are the basis for the problem modeling and policy evaluation. However, several challenging issues still remain. Among most cited issues, the enormity of state/action space is an important factor that causes inefficiency in accurately approximating the state-action-value function. We observe that although actions directly define the agents' behaviors, for many problems the next state after a state transition matters more than the action taken, in determining the return of such a state transition. In this regard, we propose a new learning paradigm, State Action Separable Reinforcement Learning (sasRL), wherein the action space is decoupled from the value function learning process for higher efficiency. Then, a light-weight transition model is learned to assist the agent to determine the action that triggers the associated state transition. In addition, our convergence analysis reveals  that under certain conditions, the convergence time of sasRL is $O(T^{1/k})$, where $T$ is the convergence time for updating the value function in the MDP-based formulation and $k$ is a weighting factor. Experiments on several gaming scenarios show that sasRL outperforms state-of-the-art MDP-based RL algorithms by up to $75\%$. 
\end{abstract}

\section{Introduction}

Classic Reinforcement Learning (RL) \cite{suttonbarto98} methods, which were developed to solve serial decision-making and control problems,  have been investigated for decades. For instance, $Q$-learning algorithm \cite{watkins1992q}, which first appeared in the late 1980s and had since been thoroughly studied and analyzed, inspires many successful algorithms and applications. However, due to the lack of general means for function approximation, value functions in $Q$-learning were estimated in tabular settings or by using simple linear parametrizations. As such, their applicabilities are limitted to some simple problems with relatively small state-action spaces. In recent years, the advancements in deep learning \cite{bengio2009learning} extend RL to Deep Reinforcement Learning (DRL) \cite{mnih2013playing}, for which Deep Neural Networks (DNN) \cite{hagan1995neural} are employed as value function approximators. 

Conventionally, Markov Decision Process (MDP) is used to model RL problems. For an MDP, the RL agent jumps between states by taking actions and it collects a reward after transitioning from one state to the next state. The agent maintains a state-action-value (SAV) function (e.g., the $Q$-value \cite{watkins1992q}) to estimate the long-term returns of state action pairs. This SAV function is iteratively updated using rewards associated with state transitions. For DRL methods based on MDPs, both model-based and model-free \cite{suttonbarto98} approaches use DNNs as approximators for SAV functions. For example,  Google's DQN \cite{mnih2015human} and a family of algorithms using the actor-critic framework \cite{konda2000actor} all share this common feature, despite some variations in design details. Then, the agent's policy is developed, either directly or indirectly, according to the learned SAV function.  

The SAV function based on the MDP formulation is a convenient choice for developing policies, since actions directly define the behaviors of RL agents. By coupling its behaviors with potential returns, one implicit assumption is that the long term return of an agent is a function of its current state and available actions. Although this is generally true, for many RL problems, the return of a state transition is directly determined by the next state after the state transition, and the action is only indirectly related to the return as it causes such a state transition. 
Furthermore, for some RL problems, there are potentially several actions that can cause the same state transition. Then, all these actions have the same effect as far as rewards are concerned. As a result, this induces extra burdens for training the DNN function approximators. In other cases, due to the stochastic nature of the environment, the same action can cause different state transitions with distinct rewards from the common current state. This can be troublesome for the DNN function approximator of the SAV function, since the same input (the state action pair) is trained to produce different outputs.
All these factors considered, we argue that although SAV function based on MDP can be intuitive and convenient, the aforementioned issues can cause difficulties and inefficiencies in the training process. 

Aimed to address these issues for RL tasks where rewards are tightly associated with state transitions, we propose an alternative RL paradigm, called State Action Separable Reinforcement Learning (sasRL), by formulating the RL problem as a modified Markov Reward Process (mMRP, defined in Section~\ref{sec:mMRP}). Specifically, we employ a new value function, the \emph{state-transition-value (STV)}, to estimate returns of state transitions. The STV function takes the current and the next state as input and estimates the return of such a state transiton pair. While the STV function is targeted at addressing the issues discussed above, another added benefit is that in this way, the input to the DNN function approximator only spans the state space. This is in comparison to the case of SAV function used for MDP, whose DNN approximator takes inputs that span both state and action spaces. Our intuition is that the input dimensionality reduction speeds up the training procedure for the DNN approximator, as the agent's actions are not explicitly modeled in mMRP.  
Therefore, sasRL develops raw policies in forms of desired next state given the current state. This is not a problem for RL tasks where the agent can determine the action to take given the desired next state. For tasks where such mappings are not obvious, we use a light-weight determinsitic transition model to help the  agent determine the action that causes the desired state transition. This transition model is trained on the same data collected for the sasRL training via standard supervised learning procedures. 
In sum, sasRL separates the RL problem into a less complicated model-free RL problem and a simple supervised learning problem. Our view is that such decoupling procedure is the key factor that leads to higher RL training efficiency and better performance. 

\section{Problem Formulation}
In this section, we introduce how an RL problem is formulated under the sasRL framework. In particular, sasRL uses modified Markov Reward Process (mMRP) to model the RL problem (Section~\ref{sec:mMRP}). Then, Section~\ref{sec:STV_function} defines the STV function under a policy for sasRL. To update the STV function and the policy, we employ the policy-gradient based method, which is described in Section~\ref{sec:policy_gradient_update}. Finally, Section~\ref{sec:transition_model} discusses how the light-weight transition model in sasRL is trained.

\subsection{The modified Markov Reward Process (mMRP)}
\label{sec:mMRP}
We propose to use a modified Markov Reward Process to model the RL problem. The Markov Reward Process (MRP) can be regarded as a Markov chain with state values added. Formally, an MRP is defined by a 4-tuple $(\mathcal{S},\mathcal{P},\mathcal{R},\gamma)$ as follows. $\mathcal{S}$ is state set.  $\mathcal{P}$ specifies the state transition probabilities, i.e., $\mathcal{P}_{ss'}=P[s_{t+1}=s'|s_{t}=s]$, where $s,s'\in \mathcal{S}$. $\mathcal{R}$ defines the rewards of states, $\mathcal{R}_{s}=\mathbb{E}[r_{t+1}|s_{t}=s]$, where $r$ is the reward. Finally, $\gamma$ is the discount factor.

For mMRP, we adopt a different reward definition while keeping everything else in the MRP unchanged. In particular, the reward function in mMRP is defined as $\mathcal{R}$: $\mathcal{R}_{ss'}=\mathbb{E}[r_{t+1}|s_{t}=s,s_{t+1}=s']$.
Note that the main difference is that for MRP, the reward only depends on the current state; while for our formulation, the reward depends on both the current and the next state of a state transition. 
Furthermore, if two actions $a_{1}$ and $a_{2}$ cause the same state transition $(s\rightarrow s')$, it is assumed that $\mathbb{E}[r(s,a_{1})]=\mathbb{E}[r(s,a_{2})]$, where $r(s,a)$ is the reward for the state action pair $(s,a)$. 

Under mMRP, the policy of an RL agent specifies the next state ($s'$) given the current state $s$. We assume that the policy is deterministic, denoted by $\mu(s)$, 
i.e., $\mu: \mathcal{S} \rightarrow\mathcal{S}$. Let the value function
under policy $\mu$ be $V^{\mu}(s)$, defined by $V^{\mu}(s)=\mathbb{E}[r_{1}+\gamma r_{2}+\dots|s_{0}=s,\mu]$, where $r_{t}\in \mathbb{R}$ is the reward at time $t$, $s_{0}$ is the initial state. The objective is to find $\mu$ that maximizes the value function.



\subsection{The state-transition-value (STV) function under the given policy}
\label{sec:STV_function}

We denote the \emph{state-transition-value (STV) function}, which quantifies the long-term return of a state transition $(s\rightarrow s')$ under policy $\mu$ as follows,
\begin{equation}
\label{eq:s_s_v}
\Phi^{\mu}(s,s')=r_{s,s'}+ \gamma \Phi^{\mu}(s',\mu(s')),
\end{equation}
where $r_{s,s'}$ is the reward for transition from $s$ to $s'$. By definition, $V^{\mu}(s) = \Phi^{\mu}(s,\mu(s))$. 
Then, the return of the policy $\mu$, $J(\mu)$ , which corresponds to the optimization goal, is 
\begin{equation}
\label{eq:policy_objective}
J(\mu)=\int_{s\in \mathcal{S}}\rho^{\mu}(s)V^{\mu}(s)\text{d}s, 
\end{equation} 
where $\rho^{\mu}(s)$ is the discounted state distribution under $\mu$ \cite{silver2014deterministic,1906.07073}. Let the initial state and the initial state distribution be $s_{0}$ and $p_{0}(s_{0})$, respectively. Then, $\rho^{\mu}(s) = \int_{s_{0} \in \mathcal{S}}\Sigma_{t=0}^{\infty}\gamma^{t}p_{0}(s_{0})p(s_{0}\rightarrow s, t,\mu)\text{d}s_{0}$, where $p(s_{0}\rightarrow s, t,\mu)$ denotes the probability of transitioning from state $s_{0}$ to state $s$ after $t$ steps.

\subsection{Policy-gradient based learning}
\label{sec:policy_gradient_update}
The STV function defined in (\ref{eq:s_s_v}) is the basis for deriving policies for the RL problem. In order to obtain an accurate STV function, it is iteratively updated using $(s,s',r)$ tuples. 
A more intuitive way is to parametrize the policy and update its parameters, so that policies can be directly generated. This approach is referred to as policy-gradient method \cite{sutton2000comparing}. For generality, we assume that STV function and policy are parametrized by parameter set $\boldsymbol{\kappa}$ and $\boldsymbol{\theta}$, denoted by $\Phi_{\boldsymbol{\kappa}}$ and $\mu_{\boldsymbol{\theta}}$, respectively. For brevity, we use $\Phi_{\boldsymbol{\kappa}}$ and $\Phi$, $\mu_{\boldsymbol{\theta}}$ and $\mu$ interchangeably. Next, we discuss how $\Phi_{\boldsymbol{\kappa}}$ and $\mu_{\boldsymbol{\theta}}$ are updated.

First, the STV function parameter set $\boldsymbol{\kappa}$ is updated by minimizing the mean squared TD(0) error \cite{suttonbarto98} defined as $L=\big(r_{s,s'} + \gamma\Phi^{\mu}_{\boldsymbol{\kappa}}(s',\mu(s')) - \Phi^{\mu}_{\boldsymbol{\kappa}}(s,s')\big)^{2}$.
Second, recall that the return of the policy $\mu$ is defined as $J(\mu)$ in (\ref{eq:policy_objective}). Therefore, the policy-gradient method aims to maximize $J(\mu_{\boldsymbol{\theta}})$ by performing gradient ascent on parameter set $\boldsymbol{\theta}$ using the policy gradient $\nabla_{\boldsymbol{\theta}}J(\mu_{\boldsymbol{\theta}})$.  In the following theorem, we present how such policy gradient is computed. 


\begin{theorem}
\label{the:policy_gradient_theorem}
If $V^{\mu_{\boldsymbol{\theta}}}(s)$ and $\nabla_{\boldsymbol{\theta}}V^{\mu_{\boldsymbol{\theta}}}(s)$ are continuous function of $\boldsymbol{\theta}$ and $s$, then the following holds,
\begin{equation} \label{eq3}
        \begin{split}
        \nabla_{\boldsymbol{\theta}}J(\mu_{\boldsymbol{\theta}}) 
        & \approx \int_{s\in\mathcal{S}}\rho^{\beta}(s)\nabla_{\boldsymbol{\theta}}\mu_{\boldsymbol{\theta}}(s)\nabla_{s'}\Phi^{\mu_{\boldsymbol{\theta}}}(s,s')\text{d}s \\
            & = \mathbb{E}_{s\sim\rho^{\beta}}\Big[\nabla_{\boldsymbol{\theta}}\mu_{\boldsymbol{\theta}}(s)\nabla_{s'}\Phi^{\mu_{\boldsymbol{\theta}}}(s,s')|_{s'=\mu_{\boldsymbol{\theta}}(s)}\Big],                    
        \end{split}
    \end{equation}
\end{theorem} 

where $\beta$ denotes the behavior policy \cite{degris2012off} used to generate training data, and $\rho^{\beta}(s)$ is the discounted distribution of states under the behavior policy. 

Let $\alpha_{\boldsymbol{\theta}}$ be the learning rate for updating $\boldsymbol{\theta}$. Then, the policy parameter is updated as follows,
    \begin{equation}
    \label{eq:policy_update}
        \boldsymbol{\theta} \leftarrow \boldsymbol{\theta} + \alpha_{\boldsymbol{\theta}}\nabla_{\boldsymbol{\theta}}\mu_{\boldsymbol{\theta}}(s)\nabla_{s'}\Phi^{\mu_{\boldsymbol{\theta}}}(s,s').
    \end{equation} 

\begin{figure}
	\smallskip
	\centering
	\begin{subfigure}[b]{0.325\textwidth}
		\centering
		\includegraphics[width=\linewidth,height=0.6\linewidth]{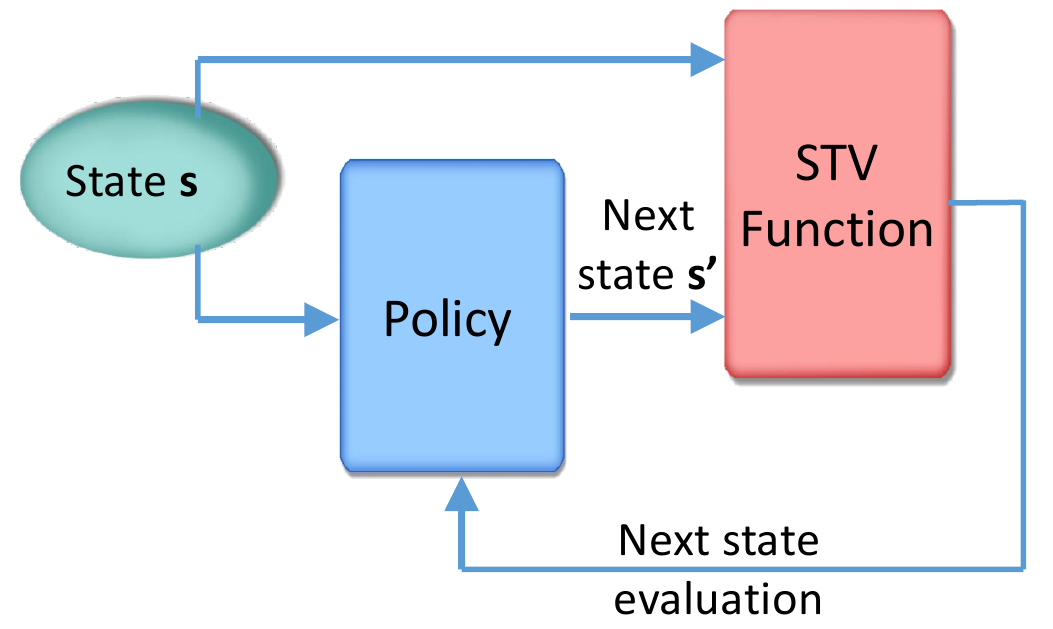}
		\caption{Policy and STV function updates by policy-gradient methods.}
		\label{fig:f0_1}
	\end{subfigure}
	\begin{subfigure}[b]{0.325\textwidth}
		\centering	\includegraphics[width=0.7\linewidth,height=0.6\linewidth]{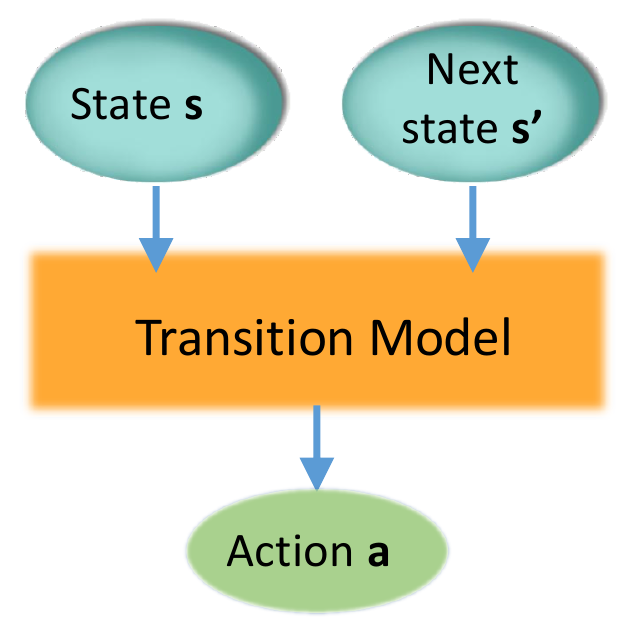}
		\caption{State transition model training by supervised learning.}
		\label{fig:f0_2}
	\end{subfigure}   
	\begin{subfigure}[b]{0.325\textwidth}
		\centering		\includegraphics[width=0.7\linewidth,height=0.6\linewidth]{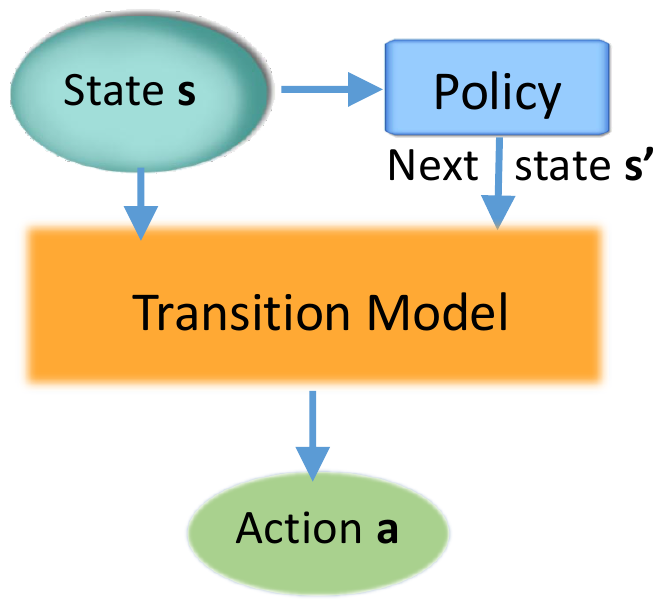}
		\caption{sasRL in operation with trained policy and transition model.}
		\label{fig:f0_3}
	\end{subfigure}
	\caption{The training and operation of sasRL.}
\label{fig:f1}
\end{figure}

Figure~\ref{fig:f0_1} demonstrates the policy-gradient based updates for the policy and STV function in sasRL.

\subsection{Deterministic state transition model and its training}
\label{sec:transition_model}
The mMRP formulation for sasRL does not explicitly model actions of the agents, and the policy developed based upon it indicates the target next state ($s'$) given the current state ($s$). For RL tasks with state space consisting of hand-crafted features, it can be straightforward for the agent to determine the action that causes the state transition $(s\rightarrow s')$. 
In other RL tasks where action cannot be determined from $(s\rightarrow s')$, we build a light-weight deterministic transition model to help the agent determine the action that can cause the state transition $(s\rightarrow s')$. This deterministic transition model can be represented by a DNN whose parameters are optimized using standard supervised learning techniques. Formally, define the deterministic transition model $\tau_{\boldsymbol{\omega}}:\mathcal{S}\times\mathcal{S}\rightarrow\mathcal{A}$, which is parametrized by a set of DNN weights $\boldsymbol{\omega}$. Then, we train the  model by minimizing the prediction error, $\mathcal{L}_{\boldsymbol{\omega}}$, by using the samples $(s,s',a,r)$ which are collected for the RL training. In particular, the loss $\mathcal{L}_{\boldsymbol{\omega}}$ is defined as $\mathcal{L}_{\boldsymbol{\omega}} = L (\tau_{\boldsymbol{\omega}}(s,s'),a)$,
where the type of loss $L$ depends on the representation of the action vector (e.g., $L$ uses binary cross entropy loss if the action vector consists of only $0$ and $1$ elements). Figure~\ref{fig:f0_2} shows the supervised learning process for training the transition model. After all components of sasRL, i.e., the parametrized policy, the parametrized STV function, and the transition model, are trained, Figure~\ref{fig:f0_3} describes sasRL in operation. 

\section{Convergence Analysis}
We conduct convergence analysis for sasRL based on existing convergence studies developed for the classic $Q$-learning algorithm. 
Due to page limit, we present our main results in this section and document related theorems and lemmas in the appendix. 

Let $\mathcal{S}$ and $\mathcal{A}$ be the state and action spaces of the RL problem under the MDP formulation. Then, let $\beta(s)$ be the given sampling policy which generates data for updating value function and policy parameters. At each time step, an action is chosen according to $\beta$ and the state transition $s+a\rightarrow s'$ takes place. Let $W$ be the total time steps for one data generation trajectory. All $W$ state transitions are recorded in forms of $W$ $(s,a)$ and $(s,s')$ pairs. Denote by $\nu(s,a)$ and $\nu(s,s')$  the number of occurrence of $(s,a)$ and $(s,s')$ pairs, respectively. Then, $p^{\beta}(s,a) = \mathbb{E}[\nu(s,a)/W]$ and $p^{\beta}(s,s') = \mathbb{E}[\nu(s,s')/W]$ are the the probabilities of recording $(s,a)$ pairs under the sampling policy $\beta$. In addition, let $p^{\beta}_{\min/\max}(s,a)$ and $p^{\beta}_{\min/\max}(s,s')$ be the  minimum/maximum $p^{\beta}(s,a)$ and $p^{\beta}(s,s')$, respectively. We define  $R_{1} = p^{\beta}_{\min}(s,a)/p^{\beta}_{\max}(s,a)$ and $R_{2} = p^{\beta}_{\min}(s,s')/p^{\beta}_{\max}(s,s')$. Then, we summarize the convergence comparison results for SAV and STV functions as follows.

\begin{proposition}
\label{prop:1}
Given the relevant conditions discussed in the appendix are met, it holds that the convergence time for updating the STV function when the RL problem is formulated by the mMRP framework is $O(T^{1/k})$, where $k = R_{2}/R_{1}$, and $T$ is the convergence time for updating the SAV function when the RL problem is formulated and trained under the MDP framework.  
\end{proposition}

\emph{\textbf{Efficient Training Condition ($k> 1$):}} Proposition~\ref{prop:1} reveals the key threshold of $k=1$, i.e., when $k>1$, the value function update convergence under the mMRP formulation is faster than that under the MDP formulation for RL problems. Recall that $k$ is closely related to the behavior (sampling) policy, which is used to collect data for off-policy\cite{degris2012off} updates of the value functions. In this paper, we argue that sasRL is most suitable for those problems where the action space is large and multiple actions can trigger the same or similar state transitions. Indeed, many RL problems of such a nature result in $k> 1$ under the given behavior policies (see Section~\ref{sec:scenario_discussion} for more discussions). 

\section{The Embodiment of sasRL}
To demonstrate how sasRL can be implemented in practice, here we implement an instance of sasRL using the actor-critic framework \cite{konda2000actor}, which is suitable for the policy-gradient based update process described in Section~\ref{sec:policy_gradient_update}. The actor-critic framework offers a natural way to concurrently optimize policy parameters and the STV function parameters. There are several advantages for adopting an actor-critic approach, compared to  more straightforward methods such as the Monte-Carlo REINFORCE \cite{williams1992simple} algorithm. The most obvious one is that actor-critic methods are intuitive as policies can be directly derived using the trained DNN with policy parameters, which is especially useful for RL tasks with large action spaces. 
Moreover, since actor-critic methods do not require whole trajectories, they can be implemented online or for non-episodic problems.

\begin{algorithm}[tb]
	\caption{sasRL training procedure}
	\label{alg:aglorithm1}	
	\Indp // \texttt{policy training} \\
	Initialize STV function (critic) parameters $\boldsymbol{\kappa}$ and policy (actor) parameters $\boldsymbol{\theta}$; \\
	Initialize delayed parameters $\boldsymbol{\kappa}'\leftarrow\boldsymbol{\kappa}$, and $\boldsymbol{\theta}'\leftarrow\boldsymbol{\theta}$;\\
	Initialize replay buffer with $(s,s',a,r)$ tuples generated by behavior policy $\beta$. \\

\While{the maximum number of iterations not reached OR not converged}
	{	
	Pull a random minibatch of $(s,s',a,r)$ from the replay buffer;	
		
		Update the critic parameters $\boldsymbol{\kappa}$ by minimizing the mean squared TD(0) error: $L = \big(r_{s,s'} + \gamma\Phi^{\mu_{\boldsymbol{\theta}'}}_{\boldsymbol{\kappa}'}(s',\mu_{\boldsymbol{\theta}'}(s')) - \Phi^{\mu_{\boldsymbol{\theta}}}_{\boldsymbol{\kappa}}(s,s')\big)^{2}$; \\					    
			Update the actor parameters $\boldsymbol{\theta}$ according to (\ref{eq:policy_update});\\
			Soft update the delayed parameters: $\boldsymbol{\kappa}'\leftarrow \epsilon \boldsymbol{\kappa} + (1-\epsilon)\boldsymbol{\kappa}'$, $\boldsymbol{\theta}'\leftarrow \epsilon \boldsymbol{\theta} + (1-\epsilon)\boldsymbol{\theta}'$;\\
			\If{current policy is evaluated}
			{  
			   store new $(s,s',a,r)$ samples collected from roll-out episodes in replay buffer. \\			
			}
	}
// \texttt{transition model training (optional) using data from the replay buffer} \\
\While{the transition model training not converged}
	{
Pre-process (see the appendix for details) minibatch of data for training the transition model;\\
Update transition model parameter $\boldsymbol{\omega}$ by minimizing loss $L (\tau_{\boldsymbol{\omega}}(s,s'),a)$.
	}
\end{algorithm}

The structure of the actor-critic implementation of sasRL is similar to the update process shown in Figure~\ref{fig:f0_1}. Specifically, the actor and critic correspond to the policy and STV function, respectively. During training, the actor-critic model is used to concurrently update $\boldsymbol{\kappa}$ and $\boldsymbol{\theta}$ which are the weights of the actor and the critic, respectively. The training samples generated by the behavior policy $\beta$ are organized in $(s,s',a,r)$ tuples. These tuples are stored in the replay buffer \cite{mnih2015human} to be used for training multiple times. The training procedure of sasRL is summarized in Algorithm~\ref{alg:aglorithm1}. Note that delayed parameters are used for training the critic network (Line 7 of Algorithm~\ref{alg:aglorithm1}), which is  a foundation technique \cite{mnih2015human,lillicrap2015continuous} in DRL literature to stabilize DRL training. 

The actor-critic part of sasRL is model-free, since both the actor and the critic learn directly from samples without explicitly requiring any modelings of the mMRP. The actor and critic in this sasRL embodiment are both implemented as multi-layer perceptrons (MLPs) \cite{gardner1998artificial}; their specifications are documented in the appendix. As for the optional light-weight transition model in sasRL, it is trained using $(s,s',a)$ tuples. The input to the transition model is $(s,s')$ pair and the output is the action $a$ that causes this state transition. 
The transition model is also built as an MLP. 

\section{Experiments}
The main objective of our experiments is to evaluate the performance of sasRL which is pertinent to its structure and the mMRP problem formulation. Therefore, we strive to minimize the influence of other factors such as the design of the DNN and its hyper-parameters. For these considerations, our experiment scenarios do not involve heavy imagery or high-dimensional state definitions, for which extra efforts in parameter tuning and model design are needed. Since we argue that the formulation based on mMRP is more efficient in learning the value function, we compare the performance of sasRL with state-of-the-art DRL solutions based on the MDP formulation. Due to page limit, we only provide high level descriptions of our experiments here, with further details in the appendix.  

\subsection{Baselines}
1). \emph{DDPG: Deep Deterministic Policy Gradient \cite{lillicrap2015continuous}} is a model-free and off-policy DRL algorithm based on the deterministic policy gradient theorem \cite{silver2014deterministic}. DDPG employs several techniques to improve data usage efficiency and to stabilize the DRL training process, such as replay buffers and the soft parameter update procedure. 

2). \emph{SAC: Soft Actor-Critic \cite{haarnoja2018soft}} is a model-free and off-policy RL algorithm for learning stochastic policies. SAC is partially inspired by the desire to address DDPG's brittleness and hyperparameter sensitivity. To this end, SAC maximizes the trade-off between the policy's performance and its randomness, which is measured by entropy. 

3). \emph{PPO: Proximal Policy Optimization algorithms \cite{schulman2017proximal}} is a relatively light-weight, model-free, and on-policy RL algorithm for learning stochastic policies. The core idea is to ensure the policy update does not go too far, while striving for greater improvements per update. In particular, PPO relies on the clipping of the objective function, among other techniques, to achieve this goal. Another distinctive feature of PPO is that it requires consecutive data samples (i.e., trajectories) for policy update. 

For fair and consistent comparison, we use reference implementations of these baseline algorithms from the Stable Baselines project \cite{stable-baselines}.  

\begin{figure}
	\smallskip
	\centering
	\begin{subfigure}[b]{0.325\textwidth}
		\centering
		\includegraphics[width=\linewidth,height=0.65\linewidth]{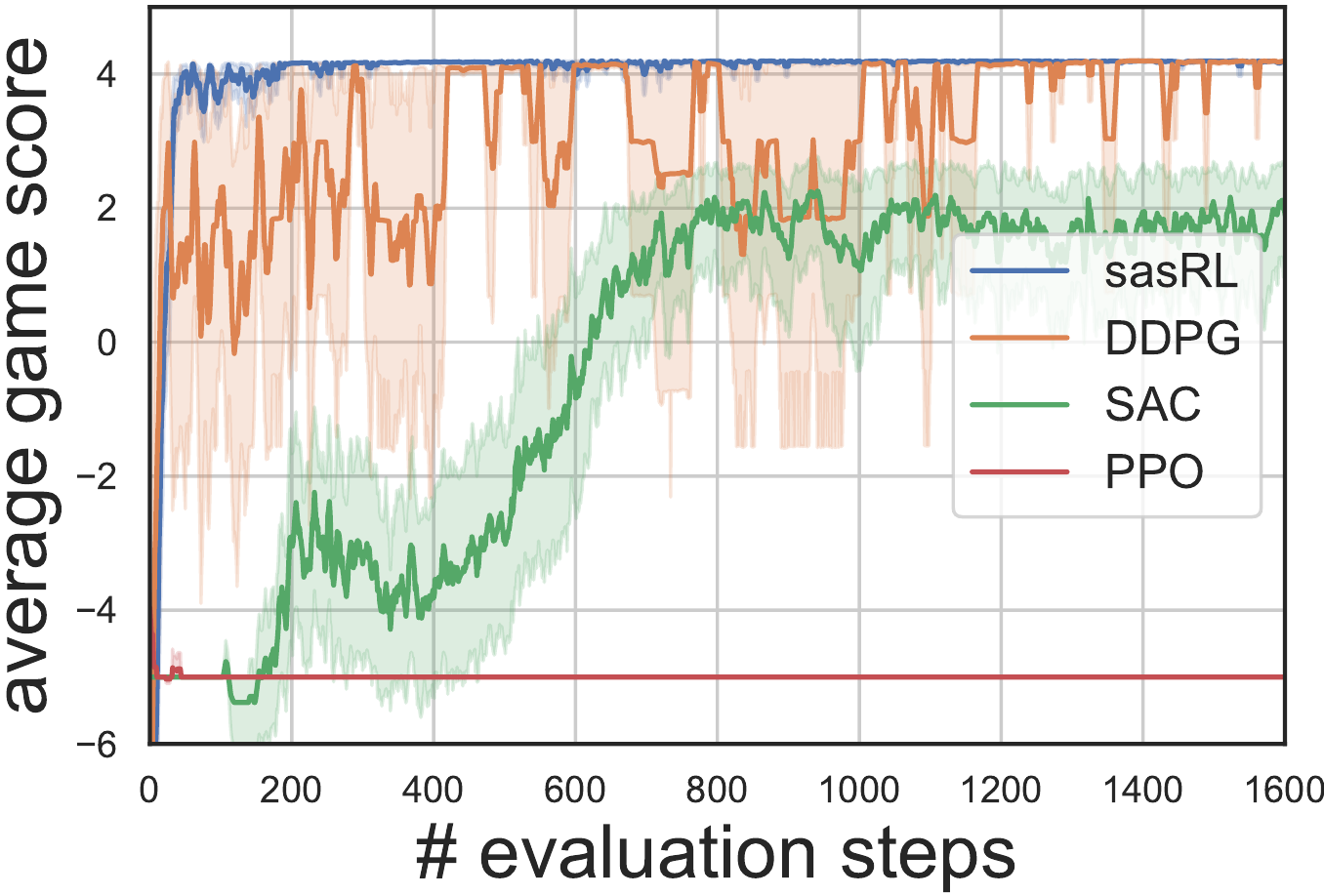}
		\caption{Grid world exit.}
		\label{fig:f1_1}
	\end{subfigure}
	\begin{subfigure}[b]{0.325\textwidth}
		\centering
		\includegraphics[width=\linewidth,height=0.65\linewidth]{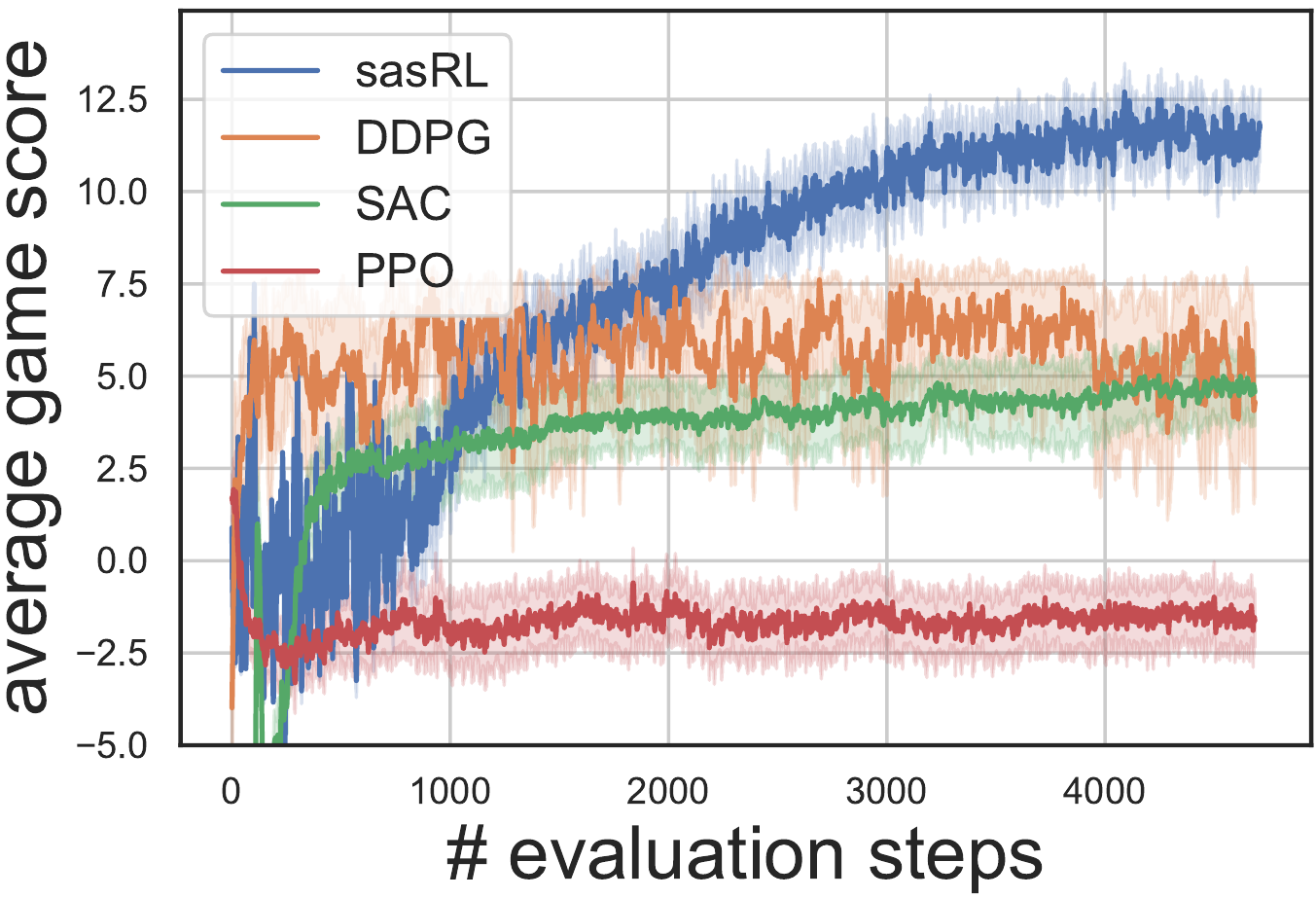}
		\caption{Berzerk.}
		\label{fig:f1_2}
	\end{subfigure}   
	\begin{subfigure}[b]{0.325\textwidth}
		\centering
		\includegraphics[width=\linewidth,height=0.65\linewidth]{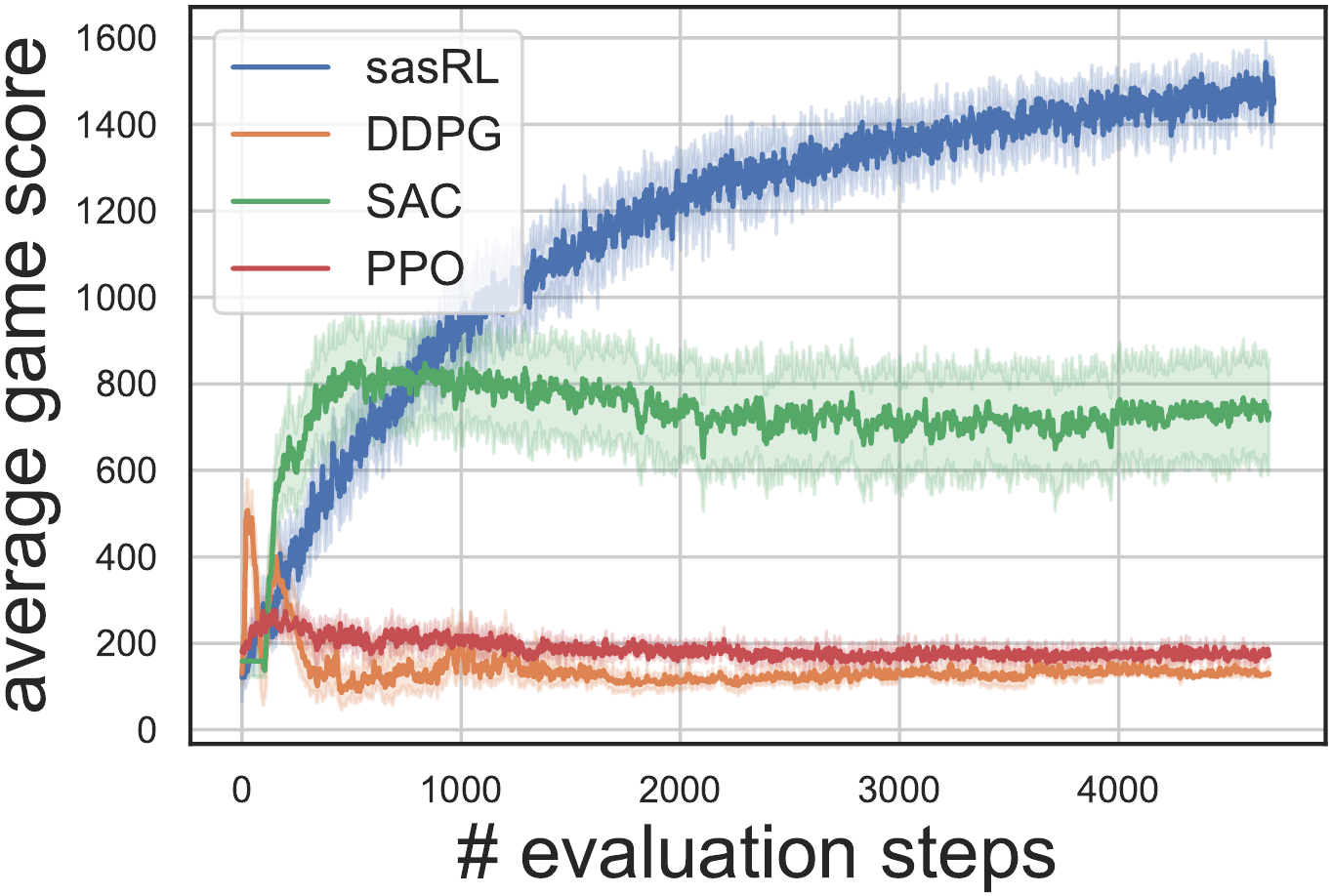}
		\caption{Slot machine.}
		\label{fig:f1_3}
	\end{subfigure}
	\caption{Comparative evaluation results.}
\label{fig:f1}
\end{figure}

\subsection{Scenarios}

1). \emph{Grid world exit problem.} First, we consider a continuous grid world exit problem where the agent tries to avoid the landmine and exit the grid as soon as possible. For this problem, the state is defined as the agent's current location. Unlike traditional grid world problem where the agent's actions are discretized as jumping from squares to squares, the scenario we consider is a continuous control problem, as the agent is allowed to move freely within certain vicinity up to a limit for each time step. The continuity in the action space increases the complexity of the SAV function and also results in potentially more actions that cause the same state transition. For each time step, the agent experiences a large negative reward for hitting a mine location, or a large positive reward for moving into the exit location. In addition, a small negative reward applies at all time steps to penalize time consumption (because the agent is expected to exit as soon as possible). 

2). \emph{Berzerk-like game.} The second scenario is a simplified berzerk game \cite{tyo2020transferable} where the agent navigates through a maze with obstacles (walls) and patrolling robots. The walls are fixed and the robots patrol on routine routes. The goal of the agent is to kill as many robots as possible by firing bullets while it tries to exit the room as soon as possible. At each time step, the agent is allowed to move freely within certain range, and one bullet is fired towards the direction of travel to kill the robot on its trajectory if any. For this problem, a state consists of the agent's location, the robots' locations, and the exit locations. The action of the agent is to move around its current location within the given limit. The reward of the agent is determined by a combination of factors detailed in the appendix.

3). \emph{The slot machine gambling game.} The third experiment considers the gambling game of a slot machine. A slot machine consists of several reels with printed symbols. The player spins the reels and receives a payout when all reels stop spinning. The payout is determined by the symbols on display on reels; see the appendix for the detailed calculation of the payout.  
Note that the player has no knowledge of how symbols are arranged on reels and cannot see the symbols before all reels stop. For this scenario, the state is defined as the symbols on display when all reels stop. For finer granularity of control, the player is allowed to decide for how long each reel spins. Therefore, the action is defined as the timer values set for all reels. The reward is defined as the payout amount. 

\subsection{Discussions on the experiment scenarios}
\label{sec:scenario_discussion}
One common feature of these experiment scenarios is that the reward of a state transition is determined by the state transition, while the action is only relevant as it causes the state transition. These are the scenarios that we argue sasRL would be more efficient than the RL algorithms based on the MDP formulation. Moreover, it is likely that multiple actions can cause the same state transition. 
As a result, our empirical results show that $k\approx 2.72$ (in Proposition~\ref{prop:1}) in this example under a random behavior policy for collecting training samples. Therefore, based on Proposition~\ref{prop:1}, faster convergence rate is expected for value function update under the mMRP. For the grid world and berzerk scenarios, due to the definition of state space, once the next state is given, the agent can directly determine the corresponding action to take. Whereas for the slot machine scenario, the agent cannot know the action that can cause a desired state transition, since the inside structure of the reel is not available to the agent. In this case, the transition model is employed to help the agent understand the state transition dynamics in relationship to actions.
Although action spaces are continuous in all scenarios, there are certain limits on each action. For example, the agent is only able to move within certain vicinity in any time step. In sasRL, to ensure that the actor generates feasible next state, its output passes through a deterministic nearest neighbor based mapping which maps the potential out-of-range next state to the feasible next state. 

\subsection{Comparative evaluation and results}

Figure~\ref{fig:f1} shows the comparisons of sasRL against DDPG, SAC, and PPO for the three evaluation scenarios. In these figures, the horizontal axis is the number of evaluation steps, while the vertical axis is the accumulated reward collected for game episodes played using the developed policy. The evaluation takes place every several gradient update steps (see the appendix for details). In our experiments, 10 instances (initialization of all DNN parameters) of these algorithms are trained and evaluated. For each evaluation episode, the maximum number of steps (cap) applies if the agent does not complete the episode when this cap is reached. In these figures, the solid lines and shaded areas are the average and the range (minimum/maximum) of the accumulated reward over all instances. 


These evaluation results show that sasRL's performance is consistently superior when compared to the baselines. In particular, PPO fails in all three scenarios. The SAC algorithm produces the most stable results on average, which is expected since SAC is designed to address the brittle convergence problem that is seen in other RL algorithms. In contrast, although DDPG outperforms SAC in grid world and berzerk scenarios, its performance is unstable, which is echoed in  \cite{haarnoja2018soft}. However, despite the inferior performance, DDPG shows fastest convergence rate in the berzerk scenario, and its convergence performance is comparable to sasRL in the grid world scenario. In summary, other than the grid world scenario where DDPG performs comparably to sasRL, sasRL outperforms all baselines for three evaluation scenarios.

\subsection{Ablation evaluation on action space granularity}
\begin{figure}
	\smallskip
	\centering
	\begin{subfigure}[b]{0.32\textwidth}
		\centering
		\includegraphics[width=\linewidth,height=0.65\linewidth]{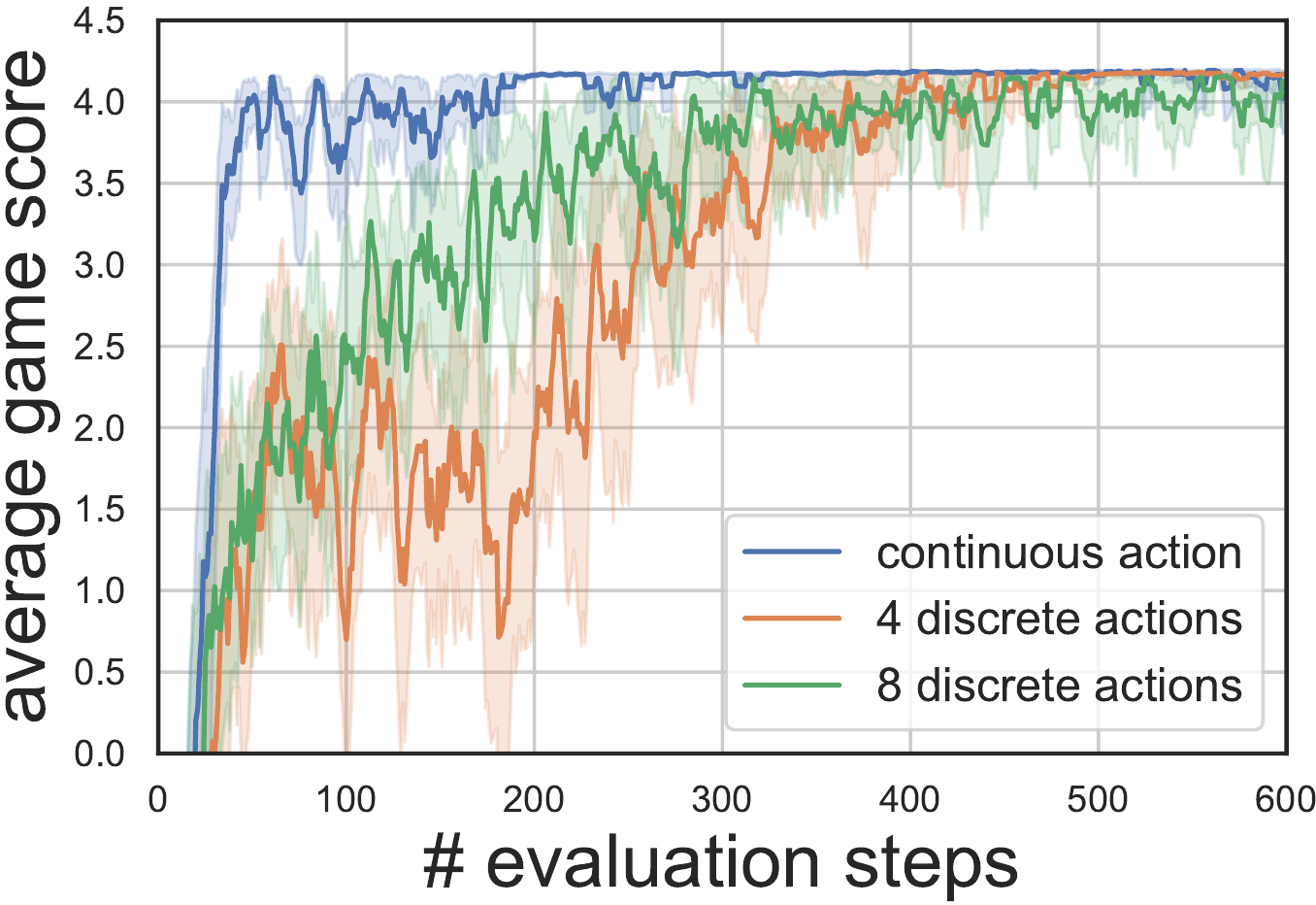}
		\caption{Grid world exit.}
		\label{fig:f2_1}
	\end{subfigure}
	\begin{subfigure}[b]{0.32\textwidth}
		\centering
		\includegraphics[width=\linewidth,height=0.65\linewidth]{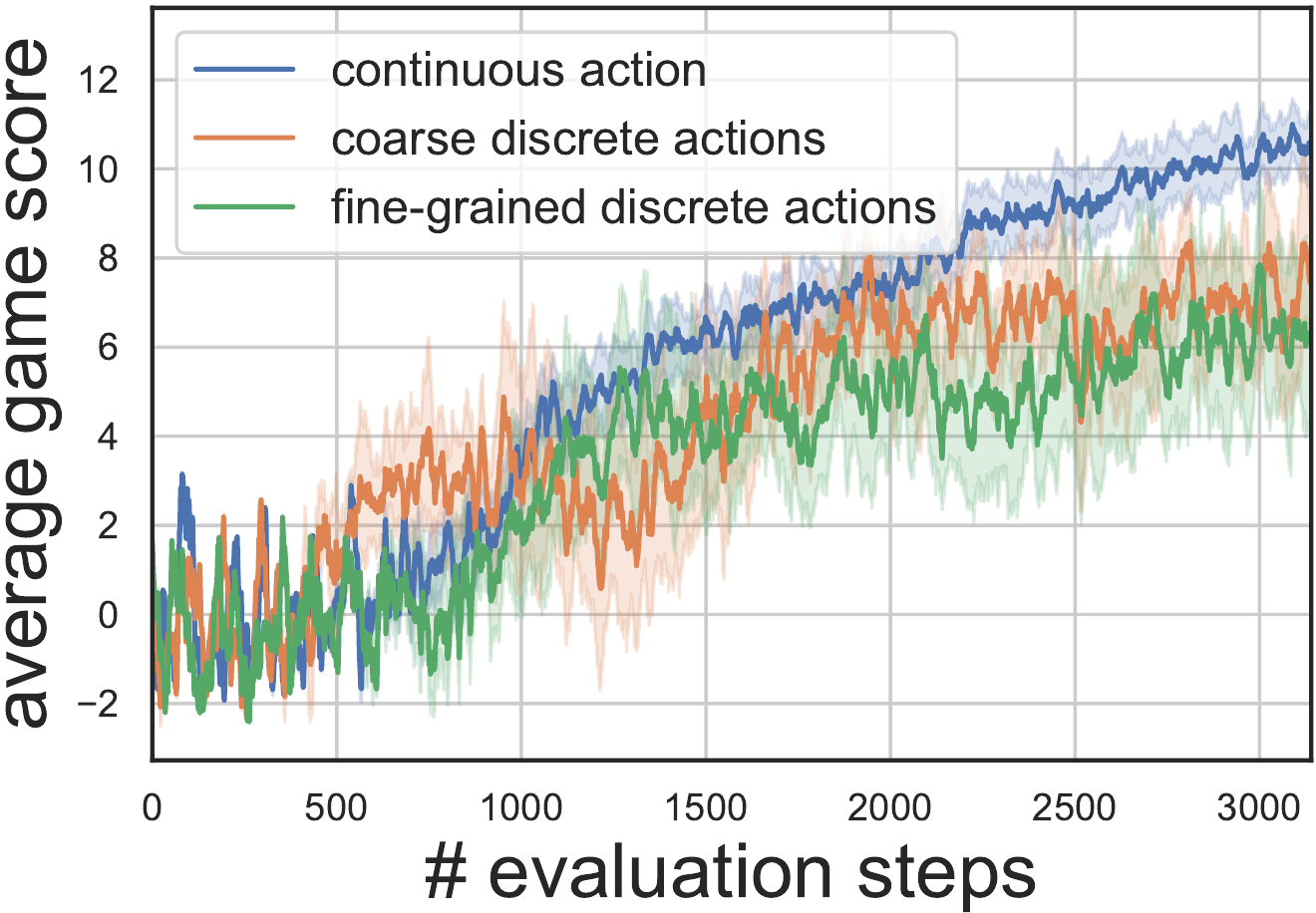}
		\caption{Berzerk.}
		\label{fig:f2_2}
	\end{subfigure}   
	\begin{subfigure}[b]{0.32\textwidth}
		\centering
		\includegraphics[width=\linewidth,height=0.65\linewidth]{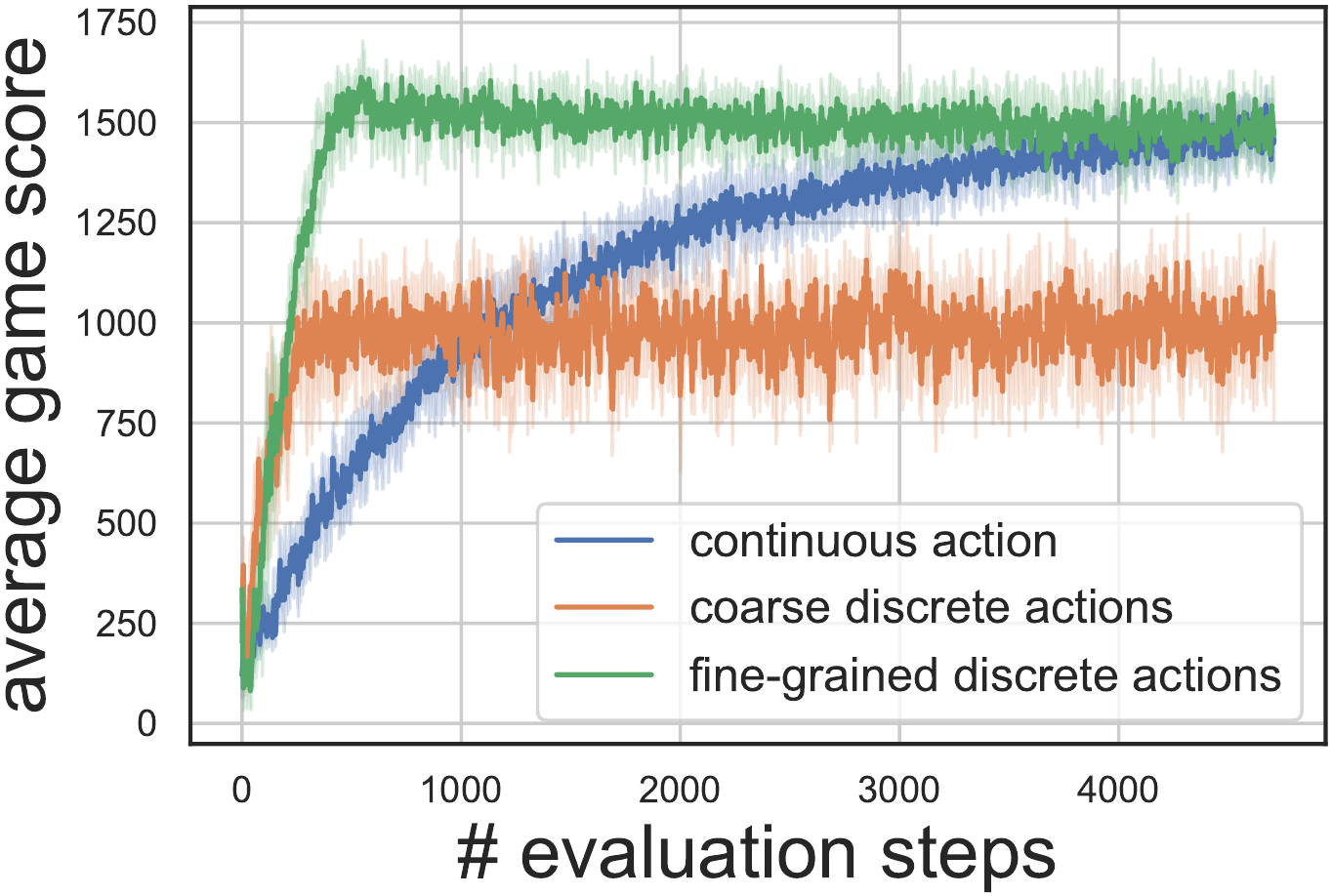}
		\caption{Slot machine.}
		\label{fig:f2_3}
	\end{subfigure}
	\caption{Evaluation results on action space granularity.}
\label{fig:f2}
\end{figure}

All evaluation scenarios we consider so far have continuous action spaces. 
Another interesting aspect to investigate is the impact of the granularity of state transitions on the performance of sasRL. To this end, we change the action space of the behavior policy from continuous to discrete for generating training samples. Specifically, we use two levels of action granularity with the discrete action space, a coarse-grained, and a fine-grained, to generate training samples. Details on the definitions for two levels of action granularity can be found in the appendix. 


The experiment results are shown in Figure~\ref{fig:f2}. Overall, they show that training samples from continuous action space enables sasRL to have better performance. This is expected as fine-grained actions result in a diverse state transition sample pools, from which the DNN function approximator can reveal more structural details of the reward dynamics. 
In the case of discrete action space, the performance of sasRL under the "coarse" and "fine-grained" action spaces are comparable, for the grid world and the berzerk scenarios. A similar trend can be observed in both scenarios that the training curves for the "coarse" cases experience more fluctuations, whereas they steadily go upwards for the "fine-grained" cases. Therefore, fine-grained state transitions tend to stabilize training. In addition, the gap between the "coarse" and "fine-grained" discrete action cases in the slot machine scenario suggests that the granularity of action space significantly influences sasRL's ability to learn for some problems. Another interesting phenomenon is that for the slot machine scenario, sasRL trained on data from continuous action space converges slower than those trained on data from discrete action spaces. Intuitively, this is caused by the transition model, which is a lot easier to train when all actions are quantized as in the case of discrete action spaces. 

\section{Related Work}
\label{sec:related_work}
sasRL is inspired by the idea of combining the strengths of both model-based and model-free RL techniques to improve training efficiency. In this regard, our work is closely related to \cite{franccois2019combined} in that \cite{franccois2019combined} trains a low-dimensional encoding of the environment, and such a encoding module is used for planning. However, the model employed in \cite{franccois2019combined} is very heavy, which requires the modeling of all elements (transition, reward dynamics, etc.) of the environment, in addition to the parametrized value function. In practice, we find that the low-dimensional encoding of the state space is rather difficult to train. Results from exisiting literatures are not conclusive on if using human insights to aid such encoding design would help DNNs better capture the state structures, as echoed by \cite{sutton2019bitter}. Similarly, other works on separating the model-free and model-based learning in RL focus on learning state, action, and/or reward representations/dynamics, separately. The central idea of these approaches is that modularizations of RL tasks have the benefit of potential transfer learning and improved learning efficiency. For example, \cite{zhang2018decoupling} decouples the RL problem into a state dynamic learning component and a reward function learning component. The learned state dynamic model is shown to be transferable to new scenarios. The method proposed in \cite{liu2019utility} offers a simple yet effective way to obtain a sparse DNN representation of the training data to assist the DRL agent in better understanding useful and pertinent dynamics in RL tasks.  On the other hand, the works in \cite{chandak2019learning,chen2019learning} investigate the embeddings of action space from theoretical and practical perspectives. Moreover, the Value Prediction Network (VPN) \cite{oh2017value} avoids the challenging task of modeling the full environment by only focusing on predicting value/reward of future states. The VPN's model-based part learns the dynamics of the abstract state transitions, while its model-free part predicts rewards and values from the abstract state space. 

Although these works use a combination of model-based and model-free RL techniques like ours, we explore the problem from different perspectives. The most distinctive difference here is that they rely on embedding and/or representation learning techniques. In contrast, we do not use any dimension reduction techniques. Instead, we decouple the action space from the model-free RL procedure and build a separate light-weight transition model which is trained via supervised learning.


\section{Conclusion}
In this paper, we presented an RL framework based on the modified MRP and the state-transition-value function. By decoupling the action space from the model-free RL procedure, we addressed the learning inefficiency issue caused by large action space and the environment stochasticity when traditional state-action value functions are used. Furthermore, a light-weight transition model is proposed to assist the agent to determine how to trigger the desired state transition when needed. For the proposed RL framework, we conducted quantitative convergence analysis to identify the conditions under which our approach converges substantially faster. Experimental results confirm the superior performance of sasRL, when compared to state-of-the-art RL algorithms developed under the MDP formulation.

\bibliography{reference.bib}



\section*{Supplementary material}

# State Action Separable Reinforcement Learning: Appendix

**Ziyao Zhang**
Imperial College London
London, United Kingdom
`ziyao.zhang15@imperial.ac.uk`

**Liang Ma**
IBM T. J. Watson Research Center
Yorktown Heights, NY, United States
`lianglondon@gmail.com`

**Kin K. Leung**
Imperial College London
London, United Kingdom
`kin.leung@imperial.ac.uk`

**Konstantinos Poularakis**
Yale University
New Haven, CT, United States
`konstantinos.poularakis@yale.edu`

**Mudhakar Srivatsa**
IBM T. J. Watson Research Center
Yorktown Heights, NY, United States
`msrivats@us.ibm.com`

## 1 Proof for Theorem 1

**Theorem 1.** *If $V^{\mu_{\boldsymbol{\theta}}}(s)$ and $\nabla_{\boldsymbol{\theta}} V^{\mu_{\boldsymbol{\theta}}}(s)$ are continuous function of $\boldsymbol{\theta}$ and $s$, then the following holds,*

$$\begin{aligned}
\nabla_{\boldsymbol{\theta}} J(\mu_{\boldsymbol{\theta}}) &\approx \int_{s \in \mathcal{S}} \rho^{\beta}(s) \nabla_{\boldsymbol{\theta}} \mu_{\boldsymbol{\theta}}(s) \nabla_{s'} \Phi^{\mu_{\boldsymbol{\theta}}}(s, s') ds \\
&= \mathbb{E}_{s \sim \rho^{\beta}} \left[ \nabla_{\boldsymbol{\theta}} \mu_{\boldsymbol{\theta}}(s) \nabla_{s'} \Phi^{\mu_{\boldsymbol{\theta}}}(s, s') |_{s' = \mu_{\boldsymbol{\theta}}(s)} \right],
\end{aligned} \quad (1)$$

where $\beta$ denotes the behavior policy [1] used to generate training data, and $\rho^{\beta}(s)$ is the discounted distribution of states under the behavior policy.

*Proof.* First, we consider the scenario of the mMRP starting from a specific initial state state $s_0$. Then, the policy gradient is calculated by $\nabla_{\boldsymbol{\theta}} J(\mu_{\boldsymbol{\theta}}) = \int_{s_0 \in \mathcal{S}} p_0(s_0) \nabla_{\boldsymbol{\theta}} V^{\mu_{\boldsymbol{\theta}}}(s_0) ds_0$, where $p_0(s_0)$ is the initial distribution of state $s_0$. Therefore, we first derive the expression of $\nabla_{\boldsymbol{\theta}} V^{\mu_{\boldsymbol{\theta}}}(s)$. Before we proceed with the derivation, we define some notations used as follows. $p(s'|s, \mu_{\boldsymbol{\theta}}(s))$ is the probability of transition into $s'$ from $s$, given the policy $\mu_{\boldsymbol{\theta}}$. Although $\mu_{\boldsymbol{\theta}}$ is a deterministic policy, the next state $s'$ is also influenced by other environment factors. For example, if the next state suggested by $\mu_{\boldsymbol{\theta}}$ is not a feasible next state, a mapping is needed to map the raw next state indicated by the

policy to a feasible next state $s'$. In addition, $p(s \to s', t, \mu_{\boldsymbol{\theta}})$ is the probability of transitioning from state $s$ to state $s'$ in $t$ steps under policy $\mu_{\boldsymbol{\theta}}$.

$$
\begin{aligned}
\nabla_{\boldsymbol{\theta}} V^{\mu_{\boldsymbol{\theta}}}(s) &= \nabla_{\boldsymbol{\theta}}(r_{s,s'} + \gamma \Phi^{\mu_{\boldsymbol{\theta}}}(s', \mu(s'))) \\
&= \nabla_{\boldsymbol{\theta}} \mu_{\boldsymbol{\theta}}(s) \nabla_{s'} r_{s,s'} + \nabla_{\boldsymbol{\theta}} \int_{s' \in \mathcal{S}} \gamma p(s'|s, \mu_{\boldsymbol{\theta}}(s)) V^{\mu_{\boldsymbol{\theta}}}(s') \mathrm{d}s' \\
&= \nabla_{\boldsymbol{\theta}} \mu_{\boldsymbol{\theta}}(s) \nabla_{s'} r_{s,s'} \\
&\quad + \int_{s' \in \mathcal{S}} \gamma \Big( p(s'|s, \mu_{\boldsymbol{\theta}}(s)) \nabla_{\boldsymbol{\theta}} V^{\mu_{\boldsymbol{\theta}}}(s') + \nabla_{\boldsymbol{\theta}} \mu_{\boldsymbol{\theta}}(s) \nabla_{s'} p(s'|s, \mu_{\boldsymbol{\theta}}(s)) V^{\mu_{\boldsymbol{\theta}}}(s') \Big) \mathrm{d}s' \\
&= \nabla_{\boldsymbol{\theta}} \mu_{\boldsymbol{\theta}}(s) \nabla_{s'} \Big( r_{s,s'} + \int_{s' \in \mathcal{S}} \gamma p(s'|s, \mu_{\boldsymbol{\theta}}(s)) V^{\mu_{\boldsymbol{\theta}}}(s') \mathrm{d}s' \Big) \\
&\quad + \int_{s' \in \mathcal{S}} \gamma p(s'|s, \mu_{\boldsymbol{\theta}}(s)) \nabla_{\boldsymbol{\theta}} V^{\mu_{\boldsymbol{\theta}}}(s') \mathrm{d}s'
\end{aligned}
\tag{2}
$$

For simpler expression, we denote $\alpha(s) = \nabla_{\boldsymbol{\theta}} \mu_{\boldsymbol{\theta}}(s) \nabla_{s'} \Big( r_{s,s'} + \int_{s' \in \mathcal{S}} \gamma p(s'|s, \mu_{\boldsymbol{\theta}}(s)) V^{\mu_{\boldsymbol{\theta}}}(s') \mathrm{d}s' \Big)$, which can be further simplified as $\alpha(s) = \nabla_{\boldsymbol{\theta}} \mu_{\boldsymbol{\theta}}(s) \nabla_{s'} \Phi^{\mu_{\boldsymbol{\theta}}}(s, s')$. Then, we have the following.

$$
\begin{aligned}
\nabla_{\boldsymbol{\theta}} V^{\mu_{\boldsymbol{\theta}}}(s) &= \alpha(s) + \int_{s' \in \mathcal{S}} \gamma p(s'|s, \mu_{\boldsymbol{\theta}}(s)) \nabla_{\boldsymbol{\theta}} V^{\mu_{\boldsymbol{\theta}}}(s') \mathrm{d}s' \\
&= \alpha(s) + \int_{s' \in \mathcal{S}} \gamma p(s \to s', 1, \mu_{\boldsymbol{\theta}}) \nabla_{\boldsymbol{\theta}} V^{\mu_{\boldsymbol{\theta}}}(s') \mathrm{d}s' \\
&= \alpha(s) + \int_{s' \in \mathcal{S}} \gamma p(s \to s', 1, \mu_{\boldsymbol{\theta}}) \Big( \alpha(s') + \int_{s'' \in \mathcal{S}} \gamma p(s' \to s'', 1, \mu_{\boldsymbol{\theta}}) \nabla_{\boldsymbol{\theta}} V^{\mu_{\boldsymbol{\theta}}}(s'') \mathrm{d}s'' \Big) \mathrm{d}s' \\
&= \alpha(s) + \int_{s' \in \mathcal{S}} \gamma p(s \to s', 1, \mu_{\boldsymbol{\theta}}) \alpha(s') \mathrm{d}s' \\
&\quad + \int_{s'' \in \mathcal{S}} \int_{s' \in \mathcal{S}} \gamma^2 p(s \to s', 1, \mu_{\boldsymbol{\theta}}) p(s' \to s'', 1, \mu_{\boldsymbol{\theta}}) \nabla_{\boldsymbol{\theta}} V^{\mu_{\boldsymbol{\theta}}}(s'') \mathrm{d}s' \mathrm{d}s'' \\
&= \alpha(s) + \int_{s' \in \mathcal{S}} \gamma p(s \to s', 1, \mu_{\boldsymbol{\theta}}) \alpha(s') \mathrm{d}s' + \int_{s'' \in \mathcal{S}} \gamma^2 p(s \to s'', 2, \mu_{\boldsymbol{\theta}}) \nabla_{\boldsymbol{\theta}} V^{\mu_{\boldsymbol{\theta}}}(s'') \mathrm{d}s'' \\
&= \vdots \\
&= \int_{x \in \mathcal{S}} \Sigma_{t=0}^{\infty} \gamma^t p(s \to x, t, \mu_{\boldsymbol{\theta}}) \alpha(s) \mathrm{d}x
\end{aligned}
\tag{3}
$$

Then, we replace $s$ with $s_0$ in $\nabla_{\boldsymbol{\theta}} J(\mu_{\boldsymbol{\theta}}) = \int_{s \in \mathcal{S}} p_0(s) \nabla_{\boldsymbol{\theta}} V^{\mu_{\boldsymbol{\theta}}}(s) \mathrm{d}s$, and use the result in (3), we obtain

$$
\begin{aligned}
\nabla_{\boldsymbol{\theta}} J(\mu_{\boldsymbol{\theta}}) &= \int_{s_0 \in \mathcal{S}} p_0(s_0) \nabla_{\boldsymbol{\theta}} V^{\mu_{\boldsymbol{\theta}}}(s_0) \mathrm{d}s_0 \\
&= \int_{s_0 \in \mathcal{S}} \Big( \int_{s \in \mathcal{S}} \Sigma_{t=0}^{\infty} \gamma^t p_0(s_0) p(s_0 \to s, t, \mu_{\boldsymbol{\theta}}) \mathrm{d}s \Big) \alpha(s_0) \mathrm{d}s_0 \\
&= \int_{s_0 \in \mathcal{S}} \rho^{\mu_{\boldsymbol{\theta}}}(s_0) \nabla_{\boldsymbol{\theta}} \mu_{\boldsymbol{\theta}}(s_0) \nabla_{s'_0} \Phi^{\mu_{\boldsymbol{\theta}}}(s_0, s'_0) \mathrm{d}s_0 \\
&= \mathbb{E}_{s \sim \rho^{\mu_{\boldsymbol{\theta}}}} \Big[ \nabla_{\boldsymbol{\theta}} \mu_{\boldsymbol{\theta}}(s) \nabla_{s'} \Phi^{\mu_{\boldsymbol{\theta}}}(s, s') |_{s' = \mu_{\boldsymbol{\theta}}(s)} \Big].
\end{aligned}
\tag{4}
$$

Note that the results in (4) is suitable for calculating on-policy policy gradient since the states are sampled by the policy being updated, i.e., $\mu_{\boldsymbol{\theta}}$. For off-policy policy gradient scenarios, where the

policy-gradient, denoted by $\nabla_{\boldsymbol{\theta}} J_\beta(\mu_{\boldsymbol{\theta}})$, is calculated using states sampled from the behavior policy $\beta$, (4) is updated as follows.

$$\begin{aligned}
\nabla_{\boldsymbol{\theta}} J_\beta(\mu_{\boldsymbol{\theta}}) &= \int_{s \in \mathcal{S}} \rho^\beta(s) \nabla_{\boldsymbol{\theta}} V^{\mu_{\boldsymbol{\theta}}}(s) \mathrm{d}s \\
&\approx \int_{s \in \mathcal{S}} \rho^\beta(s) \nabla_{\boldsymbol{\theta}} \mu_{\boldsymbol{\theta}}(s) \nabla_{s'} \Phi^{\mu_{\boldsymbol{\theta}}}(s, s') \mathrm{d}s \\
&= \mathbb{E}_{s \sim \rho^\beta} \left[ \nabla_{\boldsymbol{\theta}} \mu_{\boldsymbol{\theta}}(s) \nabla_{s'} \Phi^{\mu_{\boldsymbol{\theta}}}(s, s')|_{s'=\mu_{\boldsymbol{\theta}}(s)} \right]
\end{aligned} \quad (5)$$

Here, the second term related to the partial derivative of in $\Phi^{\mu_{\boldsymbol{\theta}}}(s, s')$ is dropped, and thus the approximation. This approximation is first used and justified in [1], and has since been commonly used in various policy gradient derivations.

□

## 2 Proofs for Convergence Analysis Results

In the high-level, our analysis first identifies the similarity of the value function update procedure of $Q$-learning and policy-gradient based method. Then, we employ similar convergence analysis derived for $Q$-learning to compare the convergence properties of the SAV and STV functions.

### 2.1 Convergence analysis for the $Q$-learning algorithm

The convergence analysis of the $Q$-learning algorithm is provided in [2]. We base part of our analysis on the main theorem from this work, which is stated as follows.

**Theorem 2.** *Let $Q_t(s, a)$ and $Q^*(s, a)$ denote the t-th iteration of the Q-function during the update process as defined in [3], and the optimal Q-function, respectively. Assume that the conditions set out in [2] are met. Then, the following relation holds asymptotically with probability one: $|Q_t(s,a)-Q^*(s,a)| \leq \frac{B}{t^{R(1-\gamma)}}$, for some suitable constant $B > 0$ when $R(1-\gamma) < 1/2$. Here, $R = p_{\min}/p_{\max}$, where $p_{\min} = \min_{(s,a)} p(s,a)$, $p_{\max} = \max_{(s,a)} p(s,a)$, and $p(s,a)$ is the sampling probability of $(s,a)$.*

### 2.2 Convergence analysis for sasRL

In order to build our convergence analysis on Theorem 2, we first identify the conditions upon which the parameter update process in policy-gradient based methods is similar to the $Q$-function update process in $Q$-learning. Then, we can employ Theorem 2 for further analysis which compares the convergence properties of SAV and STV functions, both of which use policy-gradient methods for parameter updates. In particular, we summarize these conditions in the lemma as follows.

**Lemma 1.** *For each update of the value (SAV or STV) function in policy-gradient based methods, if the policy parameters can generate policies that maximize the corresponding value function, then the value function update process in policy-gradient based methods is comparable to that of the Q-learning. As a result, the conclusion obtained in Theorem 2 is employed to analyze the convergence properties of the value function update in policy-gradient methods.*

*Proof.* We start the proof by providing an alternative view on the $Q$-function update for the classic $Q$-learning algorithm. In classic $Q$-learning algorithm where $Q$-tables are used to keep track of $Q$-values during the update process. The $Q$-function is updated as follows,

$$Q(s, a) = (1 - \alpha)Q(s, a) + \alpha\big(r + \max_{a'} Q(s', a')\big). \quad (6)$$

We take an alternative view on the $Q$-learning update process by reforming (6), which renders,

$$Q(s, a) = Q(s, a) + \alpha\big(r + \max_{a'} Q(s', a') - Q(s, a)\big)\nabla_{(s,a)} Q(s, a). \quad (7)$$

Note that (7) is obtained if we view the values of the $Q$-table as parameters of the $Q$-function and the loss function is defined as $L = \big(r + \max_{a'} Q(s', a') - Q(s, a)\big)^2/2$. Then, the $Q$-learning update

3

procedure is essentially updating the parameters of the $Q$-function approximator ($Q$-table) to make it more accurate. Therefore, this is on a par with the value function update processes used by both STV and SAV functions. We recognize that there are differences between the approximations by the $Q$-table (in $Q$-learning) and by DNNs (parametrized STV and SAV functions). Nevertheless, based on the assumption that both the $Q$-table and parametrized STV and SAV function update processes achieve exact approximation of the corresponding value functions in the asymptotic sense, their parameter update procedures are identical. Another important characteristic in the $Q$-function update procedure is that the greedy policy is employed to estimate the $Q$-value at the next state $s'$, i.e., $a' = \mathrm{argmax}_{a'} Q(s', a')$. For the policy-gradient based updates of STV and SAV functions, the equivalent update procedure requires that the value estimation for the next state is based on the action that maximizes the value function. Therefore, Lemma 1 requires the policy's ability in generating policies that maximize the corresponding value functions. □

Lemma 1 identifies the conditions under which we can compare the convergence properties of the policy-gradient based update processes of SAV and STV functions, by using already established results. Since the condition in Lemma 1 aligns with the objective of RL, without loss of generality, we assume that this condition is met and proceed with the proof of Proposition 1 as follows.

**Proposition 1.** *If the condition in Lemma 1 and all other conditions for deriving Theorem 2 are met. The convergence time for updating STV function when the RL problem is formulated by the mMRP framework is $O(T^{1/k})$, where $k = R_2/R_1$, $T$ is the convergence time for updating the SAV function when the RL problem is formulated and trained under the MDP framework.*

*Proof.* Let $f_1(s, a)$ and $f_2(s, s')$ denote the parametrized SAV and STV function, respectively. Then by Theorem 2 and Lemma 1, $|f_1(s, a) - f_1^*(s, a)| \leq \frac{B}{t^{R_1(1-\gamma)}}$ and $|f_2(s, s') - f_2^*(s, s')| \leq \frac{B}{t^{R_2(1-\gamma)}}$ hold with probability one, where $f_1^*(s, a)$ and $f_2^*(s, s')$ are the optimal SAV and STV function, respectively. We compare the convergence properties of SAV and STV functions by comparing $\frac{B}{t^{R_1(1-\gamma)}}$ and $\frac{B}{t^{R_2(1-\gamma)}}$. In particular, since $T$ is the convergence time for updating the SAV function, we solve $\frac{B}{T^{R_1(1-\gamma)}} = \frac{B}{t^{R_2(1-\gamma)}}$ for $t$, which is defined as the convergence time for updating the STV function. This renders $t = T^{R_1/R_2}$. Therefore, the convergence time for updating STV function is $O(T^{1/k})$ where $k = R_2/R_1$. □

## 3 Experiment Details

### 3.1 Experiment scenario details

1). *Grid world exit problem.* The state vector spans two-dimensional real coordinate space, i.e., $\mathcal{S} \in \mathbb{R}^2$. All state vector elements are decimal numbers between $0$ and $1$, which are the Cartesian coordinates of the agent on the two-dimensional plane. The landmine and exit locations are small regions on the Cartesian plane with each taking a $0.1 \times 0.1$ square area. For our experiments, we implement $1$ landmine location and $1$ exit location only for faster training. It is assumed that the agent always starts from the origin of the two-dimensional Cartesian plane. The action vector is also two-dimensional with its elements specifying the amounts of movement of the agent along the horizontal and vertical directions on the Cartesian plane at a time step. Note that the range of each element of the action vector is between $-0.09$ and $0.09$. The reward of the agent consists of three parts: (i) a negative reward of $-0.1$ at each time step to penalize time consumption, as the goal of the agent is to leave find the exit as soon as possible; (ii) a negative reward of $-2$ if the agent lands in the landmine location; (iii) a positive reward of $5$ if the agent finds the exit and leaves the grid world. The maximum number of time steps per game is set to $50$. The game ends automatically if the agent finds the exit before the time step limit is reached. The agent is put back to the origin when it hits a landmine while the old time step count still applies.

2). *Berzerk-like game.* The state vector consists of 12 elements, which correspond to the coordinates of the agent, 3 patrolling robots, 1 wall, and 1 exit. In particular, the robots' patrolling routines are fixed, i.e., at each time step they move to the next locations on the pre-defined routes. The action vector includes two elements which are the amount of movements of the agent along the horizontal and vertical directions, respectively. Note that at every time step, a bullet is fired towards the agent's direction of travel. The bullet can kill 1 robot if there are any on the bullet's trajectory within the same time slot of the bullet's firing. The reward consists of 4 parts as follows: (i) a negative reward

of $-0.2$ at each time step to penalize time consumption; (ii) a negative reward of $-5$ if the agent is killed by colliding with a robot; (iii) a positive reward of 3 if a robot is killed by the bullet; (iv) a positive reward of 6 if the agent exits the map alive. The maximum number of time steps per game is set to 50. The game ends automatically if either the agent is killed by collisions with a robot or the agent leaves the current map through the exit location before the time step limit is reached.

3). *The slot machine gambling game.* The slot machine in this experiment has 10 reels and 10 symbols on each reel. Note that the same symbol can appear multiple times on a reel. The state is defined as the symbols on display when all reels stop spinning. Therefore, the state vector consists of 10 elements. The value of each state element is a decimal number between 0 and 1, which is uniquely mapped to the symbol on display. Each symbol is therefore represented by the decimal numbers within a window of size 0.1. For example, the decimal values between 0 and 0.1 represent the first symbol, values between 0.1 and 0.2 represent the second symbol, etc. The action vector has 10 elements which correspond to the amount of spinning for each reel. In particular, the value of every action element is a decimal number between $-1$ and 1, which indicates the amount of spinning for each reel. A positive value indicates spinning clockwise and a negative value the opposite. For example, if the value of an action element is 1, that means the corresponding reel spins for a full cycle and the same symbol will be on display as a result when the reel stops. The reward is associated with the type and the number of symbols on display when all reels stop spinning. In particular, each symbol is associated with a different value. For each symbol, the number of that symbol on display among all reels must reach a threshold of 3 to trigger a payout. The payout of a symbol is then calculated as the product of the number of that symbol and the value of the symbol. The reward is calculated as the sum of payouts triggered by all symbols that reach the threshold. For instance, assume that symbol $a$ carries a value of 3, symbol $b$ carries a value of 4, and symbol $c$ carries a value of 5. Then, assume that among 10 symbols on display when all reels stop, there are 5 symbol $a$, 3 symbol $b$, and 2 symbol $c$. The reward is $5 \times 3 + 3 \times 4 = 27$. For each game, the player can spin all reels 20 times.

### 3.2 Implementation details of the sasRL

The actor, the critic, and the transition model in the sasRL embodiment used in our experiment are all implemented by multilayer perceptrons (MLPs).

The actor has one input layer, two hidden layers, and one output layer. The number of neurons in both the input layer and the output layer is the same as the dimension of the state vector. There are 64 neurons in both hidden layers. The rectified linear unit (ReLU) is selected as the activation function for all neurons in hidden layers, whereas sigmoid is employed as the activation function for the output layer.

The critic has two input layers, three hidden layers, and one output layer. The input layer takes the current and the next state vectors as inputs, it then concatenate the two inputs to be fed to the hidden layers. Therefore, the number of neurons in both input layers equals the dimension of the state vector. There are 128, 64, and 32 neurons in three hidden layers, respectively. The output of the the critic corresponds to the predicted STV. Therefore, there is 1 neuron in the output layer. The rectified linear unit (ReLU) is selected as the activation function for all neurons in hidden layers, whereas no activation function is used in neuron in the output layer.

The transition model also has two input layers, two hidden layers, and one output layer. The input layer takes the current and the next state vectors as inputs. Therefore, the number of neurons in the input layer is twice the dimension of the state vector. There are 64 and 32 neurons in two hidden layers, respectively. The output of the the transition model corresponds to the predicted action that causes the input state transition. Therefore, the number of neurons in the output layer is equal to the dimension of the action vector. The rectified linear unit (ReLU) is selected as the activation function for all neurons in hidden layers, whereas the hyperbolic tangent activation (tanh) is used the neuron in the output layer. Note that for the training of the transition model, the training data are pre-processed to avoid the situations where the same state transition is caused by multiple actions, resulting in various training targets for the same input $(s, s')$. Specifically, the pre-processing procedure reserves only one action as the training target for a state transition.

We use the Keras deep learning library for building and training the MLPs, which represent the actor, the critic, and the transition model described above. In particular, we use mean square error

(mse) and the Adam optimizer for estimating and minimizing training losses. The settings for the Adam optimizer are as follows: `lr` $= 0.001$, $\beta_1 = 0.9$, $\beta_2 = 0.999$, `clipnorm` $= 1.0$. The training procedure uses stochastic gradient descents operating on minibatches of data (the minibatch size is 32) for gradient updates. The policy being trained is evaluated every 20 gradient update steps. For one evaluation, the corresponding game is played according to current policy and the score at the end of the game is recorded.

For DNN weight initializations, we use the Glorot uniform initializer (also known as Xavier uniform initializer) with default settings implemented in Keras. The soft update procedure updates the delayed version of the actor and critic parameters by $10\%$ (i.e., $\epsilon = 0.1$ in line 9 of Algorithm 1), after each update of the corresponding non-delayed parameters. The discount factor for the STV function is set to $\gamma = 0.8$.

### 3.3 Details on the ablation study for action space granularity

For the grid world scenario, the coarse action space corresponds to 4 canonical actions, i.e., up, down, left, or right, whereas the fine-grained action space adds a further 4 diagonal actions, thus resulting in 8 canonical actions. As for the berzerk scenario, the agent's horizontal and vertical movements are quantized to certain number of actions. The agent can pick action(s) from the allowed action set for moving towards horizontal and/or vertical directions at a time slot. The number of quantized actions along each direction is $5$ (or $10$) for the coarse-grained (or the fine-grained) discrete action case. Finally, for the slot machine scenario, the default continuous action space means that the player can decide freely for how long each reel spins, as long as the a given maximum limit is not exceeded. For the discrete action space, the given maximum limit is quantized to several time intervals; the agent decides the time to spin for each reel by choosing from these time intervals.

### 3.4 Details on the replay buffer

The size of the replay buffer is fixed to 50000. Initially, the reply buffer is filled with data samples generated by the random behavior policy. During training, new samples are added when all samples in the replay buffer are used for training once. For each replay buffer update, 500 existing samples (or $1\%$ of all samples) are replaced with new data generated by a mix of random policy and the policy being trained. The replay buffer is randomly shuffled every time after new samples are added to it.


\end{document}